\documentclass[pdflatex,sn-mathphys-num]{sn-jnl}

\usepackage{graphicx}%
\usepackage{multirow}%
\usepackage{amsmath,amssymb,amsfonts}%
\usepackage{amsthm}%
\usepackage{mathrsfs}%
\usepackage[title]{appendix}%
\usepackage{xcolor}%
\usepackage{textcomp}%
\usepackage{manyfoot}%
\usepackage{booktabs}%
\usepackage{algorithm}%
\usepackage{algorithmicx}%
\usepackage{algpseudocode}%
\usepackage{listings}%

\theoremstyle{thmstyleone}%
\theoremstyle{thmstyletwo}%

\theoremstyle{thmstylethree}%

\begin{document}

\title[Article Title]{A Decomposition-Driven Hybrid Framework Based on STL for Accurate Traffic Flow Forecasting}


\author[1]{\fnm{Fujiang} \sur{Yuan}}\email{yuanfujiang@ctbu.edu.cn}

\author[1]{\fnm{Yangrui} \sur{Fan}}\email{YangruiFan@163.com}
\equalcont{These authors contributed equally to this work.}
      
\author*[2]{\fnm{Xiaohuan} \sur{Bing}}\email{xiaohuan.bing@stud.uni-goettingen.de}

\author[3]{\fnm{Zhen} \sur{Tian}}\email{2620920Z@student.gla.ac.uk}

\author[4]{\fnm{Chunhong} \sur{Yuan}}\email{ChYuan@stud.kpfu.ru}

\author[1]{\fnm{Yankang} \sur{Li}}\email{liyankang0101@163.com}

\affil[1]{\orgdiv{School of Computer Science and Technology}, \orgname{Taiyuan Normal University}, \orgaddress{ \city{Jinzhong}, \postcode{030619}, \state{Shanxi}, \country{China}}}
\affil[2]{\orgdiv{Institute of Computer Science}, \orgname{University of Göttingen}, \orgaddress{ \city{Göttingen}, \postcode{37077}, \state{Niedersachsen},  \country{Germany}}}
\affil[3]{\orgdiv{James Watt School of Engineering}, \orgname{University of Glasgow}, \orgaddress{ \city{G12 8QQ}, \state{Glasgow}, \country{United Kingdom}}}
\affil[4]{\orgdiv{Laboratory of Intelligent Home Appliances}, \orgname{College of Science and Technology, Ningbo University}, \orgaddress{ \city{Ningbo}, \postcode{315300}, \state{Zhejiang}, \country{China}}}


\abstract{Accurate traffic flow forecasting is crucial for intelligent transportation systems and urban traffic management. However, existing single-model approaches often struggle to capture the complex, nonlinear, and multi-scale temporal patterns inherent in traffic flow data. This study proposes a novel decomposition-driven hybrid framework that integrates Seasonal-Trend decomposition using Loess (STL) with three complementary predictive models to enhance forecasting accuracy and robustness. The STL method first decomposes the original traffic flow time series into three distinct components: trend, seasonal, and residual. Subsequently, a Long Short-Term Memory (LSTM) network is employed to model the trend component and capture long-range temporal dependencies, an Autoregressive Integrated Moving Average (ARIMA) model is applied to the seasonal component to exploit periodic patterns, and an Extreme Gradient Boosting (XGBoost) algorithm is utilized to predict the residual component and handle nonlinear irregularities. The final forecast is obtained through multiplicative integration of the three sub-model predictions. The proposed framework is validated using 998 traffic flow records collected from a New York City intersection between November and December 2015. Experimental results demonstrate that the LSTM-ARIMA-XGBoost hybrid model significantly outperforms individual baseline models including standalone LSTM, ARIMA, XGBoost, and their variants (xLSTM, sLSTM, mLSTM) across multiple evaluation metrics, including Mean Absolute Error (MAE), Root Mean Square Error (RMSE), and coefficient of determination (R²). The decomposition strategy effectively isolates different temporal characteristics, enabling each specialized model to focus on its strength domain, thereby improving overall prediction accuracy, interpretability, and stability. This hybrid approach provides a reliable and efficient solution for real-time traffic flow forecasting in urban transportation systems.}


\maketitle

\section{Introduction}\label{sec1}
\begin{figure}[htpb]
\centering
\includegraphics[width=0.78\textwidth]{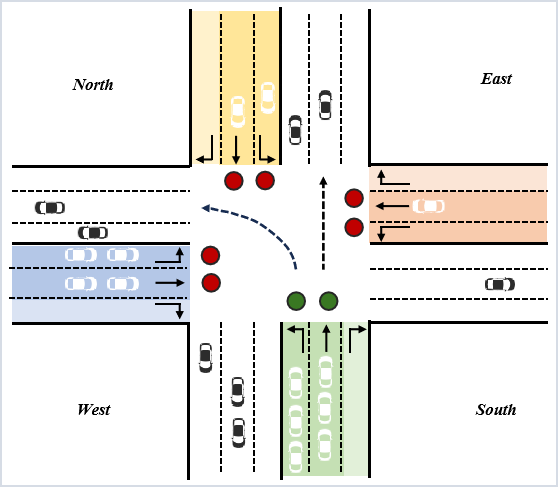}
\caption{Traffic flow distribution scenario at a typical four-way intersection on a city road}\label{fig1}
\end{figure}

In the evolution of Intelligent Transportation Systems (ITS), traffic flow prediction has played a pivotal role \cite{intro1}. Accurate and real-time traffic forecasting is not only a fundamental component of ITS but also a key enabler for efficient urban operation and intelligent mobility development \cite{intro2,zheng2025enhanced}. With the rapid increase in private vehicle ownership, particularly in fast-growing economies, urban road networks have become increasingly congested, and major intersections and arterial roads often experience persistent traffic jams \cite{intro3}. By accurately predicting traffic flow over short time intervals at critical intersections, transportation authorities can make informed decisions on traffic control and road planning, reduce accidents and delays, and provide travelers with reasonable route recommendations, thereby alleviating traffic pressure and maximizing the utilization of road resources. Figure \ref{fig1} shows the traffic flow distribution scene at a typical four-way intersection on a city road.

In traditional traffic flow prediction studies, various modeling approaches have been proposed, ranging from classical time series models (such as ARIMA) to machine learning and deep learning frameworks (such as RNN and LSTM) \cite{intro4}. Although these single-model approaches can achieve satisfactory planning performance under controlled conditions~\cite{yuan2025bio}, their generalization and robustness are often limited by the highly dynamic and nonlinear nature of urban traffic systems \cite{intro5}. Moreover, most existing models primarily emphasize prediction accuracy while overlooking critical aspects such as computational efficiency, adaptability, and scalability, which are essential for real-time applications in large-scale traffic networks  \cite{intro6}.

To address the aforementioned limitations, hybrid and decomposition-based modeling approaches have attracted growing research interest. By decomposing complex traffic flow data into more structured components and assigning suitable predictive models to each part, such frameworks can more effectively capture temporal dependencies and improve both interpretability and robustness. Building on this idea, this study introduces a hybrid forecasting framework grounded in Seasonal-Trend Decomposition using Loess. In this framework, the original traffic flow series is separated into three components: (1) a trend component predicted with a Long Short-Term Memory (LSTM) network to model long-range temporal dynamics; (2) a seasonal component captured by the Autoregressive Integrated Moving Average (ARIMA) model to reflect periodic variations; and (3) a residual component estimated using the Extreme Gradient Boosting (XGBoost) algorithm to learn nonlinear patterns and irregular fluctuations. The final forecast is obtained by integrating the outputs of these three sub-models. This decomposition-based hybrid approach effectively combines the complementary advantages of neural networks, statistical models, and ensemble learning methods. It enhances prediction accuracy, robustness, and adaptability, offering a reliable solution to the complex and dynamic challenges of traffic flow forecasting in urban transportation systems.

\section{Related Work}\label{sec2}

The literature on traffic flow forecasting reveals a broad spectrum of methodologies designed to capture the intricate spatiotemporal dependencies inherent in traffic data. Early efforts, such as those by Li et al. \cite{bib1}, introduced the Spatial–Temporal Fusion Graph Neural Network (STFGNN) to address the challenges posed by complex spatial correlations and dynamic temporal variations across road networks. This line of research underscores the necessity of jointly modeling spatial and temporal information through graph neural networks to enhance forecasting accuracy.

Building upon this spatiotemporal modeling paradigm, subsequent studies have incorporated more advanced deep learning architectures. Fang et al. \cite{bib3} proposed a model that integrates attention mechanisms with Long Short-Term Memory (LSTM) networks, demonstrating the effectiveness of attention in identifying salient temporal features. Similarly, Wang et al. \cite{bib7} developed an Attention-Based Spatiotemporal Graph Network, further reinforcing the role of attention mechanisms in capturing intricate dependencies within traffic data. Graph Neural Networks (GNNs) have emerged as a dominant framework in this domain. Zhang et al. \cite{bib4} presented a graph-based temporal attention model tailored for multi-sensor traffic flow forecasting, highlighting the advantages of leveraging sensor data within graph structures. Extending this idea, Chen et al. \cite{bib5} proposed TrafficStream, a continual-learning-based GNN framework capable of adapting to evolving traffic networks, particularly in the context of expanding sensor infrastructures. Transformer-based models have also gained considerable attention due to their strong capability in modeling long-range temporal dependencies. Reza et al. \cite{bib6} introduced a multi-head attention transformer model and provided a comparative analysis against recurrent neural networks, illustrating the transformer’s superior forecasting potential. Complementing this, Huo et al. \cite{bib9} developed a hierarchical spatiotemporal model that integrates a long-term temporal transformer with spatiotemporal graph convolutional networks, achieving robust and consistent performance across multiple datasets.
Hybrid neural architectures that combine complementary modeling paradigms have likewise been explored. Méndez et al. \cite{bib10} employed a CNN–BiLSTM hybrid model to improve long-term traffic flow prediction, emphasizing the synergy between CNNs for spatial feature extraction and BiLSTMs for temporal sequence learning. Similarly, Djenouri et al. \cite{bib8} integrated graph convolutional networks with branch-and-bound optimization techniques to further enhance prediction accuracy through algorithmic optimization.

In addition, several recent frameworks have addressed data dynamics and scalability challenges. Chen et al. \cite{bib5} proposed a streaming GNN framework with continual learning, capable of adapting to the long-term evolution of traffic networks. Xu et al. \cite{bib11} developed a dynamic graph convolutional network designed to effectively handle temporal evolution and topological changes in traffic data.
Beyond purely data-driven approaches, some studies have explored the integration of auxiliary contextual information. Shahid et al. \cite{bib2} proposed a framework that incorporates air pollution data into traffic forecasting, suggesting that environmental factors can offer valuable supplementary insights—though this direction remains relatively underexplored.

In summary, the existing body of research highlights the growing importance of advanced neural architectures—particularly graph-based, attention-enhanced, and hybrid models—in improving the accuracy, interpretability, and robustness of traffic flow forecasting. Despite the significant progress achieved by recent studies, several limitations remain in existing traffic flow forecasting approaches. Most deep learning–based models focus primarily on improving predictive accuracy but often overlook model interpretability, computational efficiency, and adaptability to evolving traffic conditions. Graph-based and attention-driven frameworks, while powerful in capturing spatial and temporal dependencies, generally require extensive parameter tuning and large-scale datasets, which may hinder real-time deployment. Transformer and hybrid models improve long-range dependency modeling but typically entail high computational complexity and memory consumption. Moreover, many studies assume stable traffic network structures and sufficient sensor coverage, failing to account for missing data, abrupt pattern shifts, or external influences such as weather and road events. Consequently, the robustness and generalization capability of these models remain limited when applied to heterogeneous and dynamically changing urban environments.

\section{Materials and Methods}\label{sec3}
This section outlines the methodological framework adopted in this study for traffic flow forecasting. The proposed approach integrates statistical decomposition with machine learning and deep learning models to capture different characteristics of traffic time series data. First, the Seasonal-Trend decomposition using Loess method is employed to decompose the original traffic flow series into three distinct components—trend, seasonal, and residual. Each component is then modeled separately using an appropriate predictive model according to its intrinsic properties. The Long Short-Term Memory (LSTM) network is utilized to forecast the trend component, effectively capturing nonlinear and long-term temporal dependencies. The Autoregressive Integrated Moving Average (ARIMA) model is applied to the seasonal component to exploit its periodic and stationary features. The Extreme Gradient Boosting (XGBoost) algorithm is employed to predict the residual component, focusing on capturing nonlinear irregularities and random fluctuations. Finally, the predictions of these three sub-models are integrated to generate the overall traffic flow forecast. The structure of this section is organized as follows: Section~\ref{subsec2} introduces the STL decomposition method; Section~\ref{subsec3} describes the LSTM network; Section~\ref{subsec4} presents the ARIMA model; and Section~\ref{subsec5} elaborates on the XGBoost model.

\subsection{STL decomposition method}\label{subsec2}
Seasonal-Trend decomposition using Loess is a robust and flexible time series analysis method proposed by R. B. Cleveland et al. in 1990 \cite{stl1}. STL uses locally weighted regression (Loess) to smooth the original series and decompose it into three main components: long-term trend, seasonal fluctuations, and residual noise. Compared with traditional methods such as classical decomposition \cite{stl2}, moving average decomposition \cite{mad}, and empirical mode decomposition (EMD) \cite{emd}, STL has significant advantages for series with strong periodicity, non-stationarity, and local volatility. Unlike classical or moving average methods that rely on fixed seasonality assumptions and are sensitive to noise, STL can dynamically adapt to changes in trends and seasonality; at the same time, compared with EMD, which often exhibits mode mixing and limited interpretability in financial environments, the additivity and structural interpretability of STL enable each component to directly reflect the laws of the time series.

\begin{figure}[h]
\centering
\includegraphics[width=0.98\textwidth]{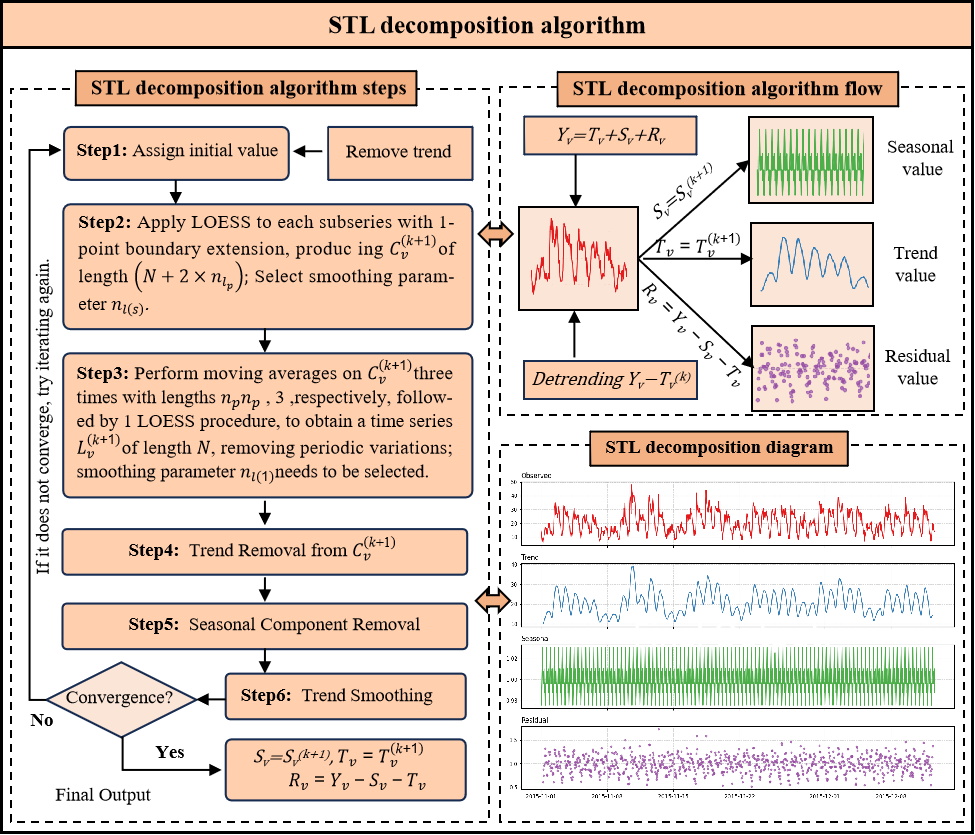}
\caption{Schematic diagram of STL decomposition algorithm}\label{fig2}
\end{figure}

The Seasonal and Trend decomposition using Loess is a robust algorithm for decomposing time series data based on locally weighted scatterplot smoothing (LOESS)  \cite{stl3}. This method decomposes the observed time series into three distinct components: trend, seasonal, and residual. The fundamental decomposition model can be expressed as:

\begin{equation}
Y_v = T_v + S_v + R_v
\label{eq:stl_decomposition}
\end{equation}

\noindent where $Y_v$ represents the original time series observations, $T_v$ denotes the trend component, $S_v$ represents the seasonal component, and $R_v$ indicates the residual component. The detailed steps of the STL decomposition algorithm are as follows:

\textbf{Step 1:} Initial Value Assignment and Trend Removal

The algorithm begins by assigning initial values to the time series and performing preliminary trend removal, yielding the detrended series:

\begin{equation}
Y_v - T_v^{(k)}
\label{eq:detrend_initial}
\end{equation}

\noindent where the superscript $(k)$ denotes the $k$-th iteration.

\textbf{Step 2:} Seasonal Component Extraction

LOESS smoothing with 1-point boundary extension is applied to each detrended subseries, generating a periodic sequence $C_v^{(k+1)}$ of length $(N + 2 \times n_{l_p})$, where the smoothing parameter $n_{l(s)}$ is selected. The mathematical expression for this step is:

\begin{equation}
S_v = S_v^{(k+1)}
\label{eq:seasonal_extraction}
\end{equation}

\noindent Through the LOESS method, the seasonal component is smoothly extracted from the detrended data.

\textbf{Step 3:} Removal of Seasonal Periodic Variations

Three moving average operations are performed on $C_v^{(k+1)}$ with window lengths of $n_p$, $n_p$, and 3, respectively, followed by a single LOESS procedure. This process yields a time series $L_v^{(k+1)}$ of length $N$, effectively removing periodic variations. The smoothing parameter $n_{l(1)}$ must be specified. This step can be expressed as:

\begin{equation}
L_v^{(k+1)} = \text{LOESS}\left(\text{MA}_{n_p}\left(\text{MA}_{n_p}\left(\text{MA}_3\left(C_v^{(k+1)}\right)\right)\right)\right)
\label{eq:moving_average}
\end{equation}

\noindent where $\text{MA}_k$ denotes the moving average operator with window size $k$.

\textbf{Step 4:} Trend Component Removal

The trend component is removed from the smoothed periodic sequence $C_v^{(k+1)}$, obtaining the detrended seasonal series:

\begin{equation}
C_v^{(k+1)} - T_v^{(k)}
\label{eq:trend_removal}
\end{equation}

\textbf{Step 5:} Seasonal Component Removal

The extracted seasonal component is removed from the original series to prepare for trend extraction:

\begin{equation}
\text{Detrending: } Y_v - S_v^{(k+1)}
\label{eq:seasonal_removal}
\end{equation}

\textbf{Step 6:} Trend Smoothing and Convergence Check

Trend smoothing operations are executed. If the algorithm converges, the final results are output; otherwise, the iteration continues. The final output is:

\begin{align}
S_v &= S_v^{(k+1)} \label{eq:final_seasonal}\\
T_v &= T_v^{(k+1)} \label{eq:final_trend}\\
R_v &= Y_v - S_v - T_v \label{eq:final_residual}
\end{align}

\subsection{LSTM Model}\label{subsec3}
Long Short-Term Memory (LSTM) networks are an improved deep learning variant of recurrent neural networks (RNNs), specifically designed to handle sequential and time-series data with long-range dependencies \cite{lstm1}. By introducing gating mechanisms—including the forget gate, input gate, and output gate—along with an internal memory cell, LSTM networks effectively regulate information transmission at each time step, thereby significantly enhancing the ability of RNNs to process long sequences \cite{lstm2}. These gates, combined with sigmoid activation functions and element-wise multiplication operations, control the intensity of information flow while preventing the accumulation of irrelevant data. The structural configuration of an LSTM network is illustrated in Figure \ref{fig2}.

\begin{figure}[H]
    \centering
    \includegraphics[width=12.5 cm]{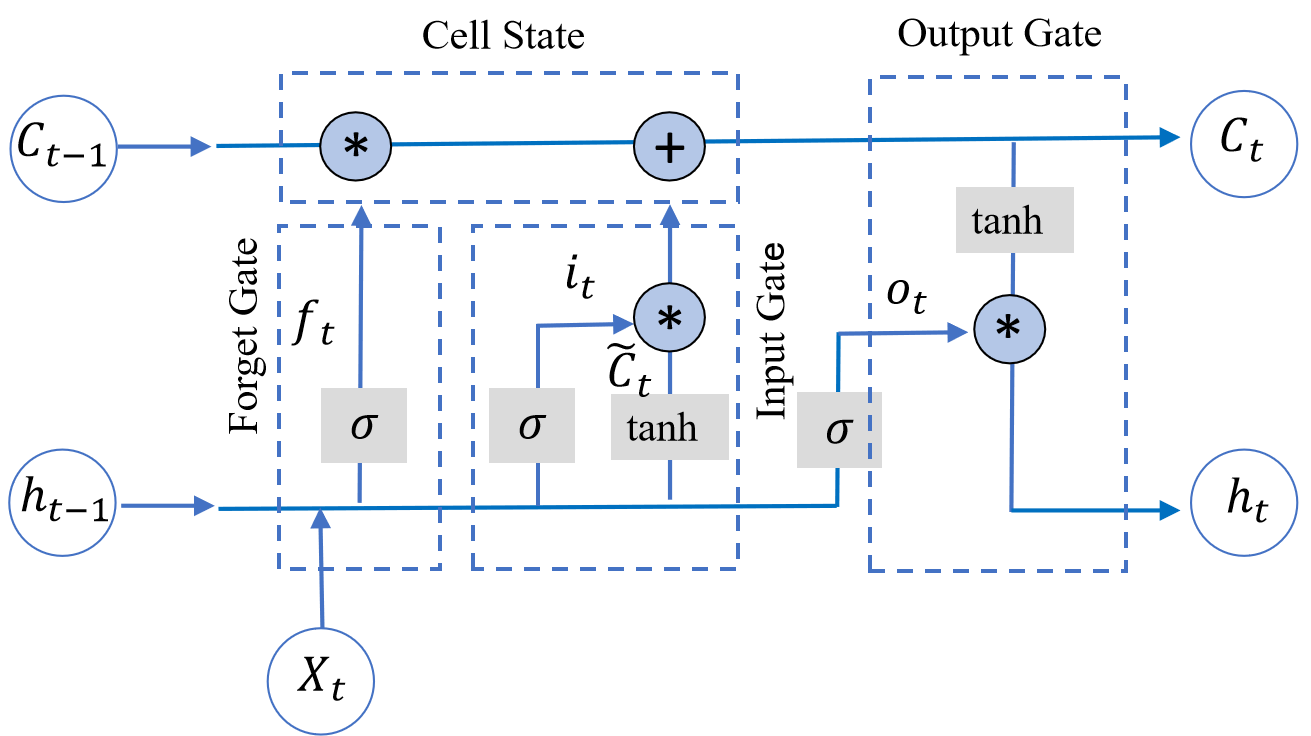}
    \caption{LSTM Architecture Diagram.}
    \label{fig3}
\end{figure}   

As illustrated in Figure \ref{fig2}, the notation $C_{t-1}$ denotes the cell state (internal memory) at the preceding time step, $X_t$ signifies the current input vector, and $h_{t-1}$ corresponds to the hidden state output from the previous time step $t-1$. At time step $t$, the forget gate generates output $f_t$, the input gate produces output $i_t$, and the output gate yields $o_t$. The sigmoid activation function is represented by $\sigma$, while $C_t$ and $h_t$ denote the current cell state and hidden state output at time $t$, respectively.

To prevent the accumulation of obsolete information that could hinder the processing of current inputs, the LSTM network employs a selective forgetting mechanism designed to discard irrelevant components from previous cell states while retaining information essential for the present computation. The forget gate regulates the proportion of historical information preserved from the cell state at time step $t-1$. Specifically, it computes a weighting factor by concatenating the current input $x_t$ with the previous hidden state $h_{t-1}$, followed by the application of a sigmoid activation function. The resulting output is a vector with elements ranging between 0 and 1, where values close to 0 denote information to be discarded, and values approaching 1 indicate information to be retained. The mathematical expression of the forget gate is formulated as follows:

\begin{equation}
    f_t = \sigma \left( W_f \cdot [h_{t-1}, X_t] + b_f \right)
\label{eq7}
\end{equation}

In Equation \ref{eq7}, the parameters $W_f$ and $b_f$ represent the weight matrix and bias vector of the forget gate, respectively, while $\sigma$ denotes the sigmoid activation function. The input gate regulates the extent to which current input information is incorporated into the cell state at the present time step, thereby managing which information requires updating. The input vector $X_t$ and previous hidden state $h_{t-1}$ are processed through the input gate, and subsequently combined with values transformed by a $\tanh$ activation function to produce updated control parameters. The mathematical representation of the input gate mechanism is given by:
\begin{equation}
    i_t = \sigma (W_i \cdot [h_{t-1}, X_t] + b_i) 
\label{eq8}
\end{equation}

\begin{equation}
    \tilde{C}_t = \tanh (W_c \cdot [h_{t-1}, X_t] + b_c) 
\label{eq9}
\end{equation}

\begin{equation}
    C_t = f_t * C_{t-1} + i_t * \tilde{C}_t 
\label{eq10}
\end{equation}

In Equation \ref{eq10}, the cell state undergoes an update to become $C_t$, where $W_i$ corresponds to the weight matrix of the input gate. The output gate controls which portion of the current cell state information is propagated to the hidden state $h_t$. Both $X_t$ and $h_{t-1}$ initially traverse the output gate to delineate the scope of information to be output. Subsequently, through integration with the $\tanh$ activation function, a selected subset of memory information from $C_t$ is processed, ultimately determining the final hidden state output $h_t$. The mathematical expression characterizing the output gate is formulated as follows:

\begin{equation}
o_t = \sigma (W_o \cdot [h_{t-1}, X_t] + b_o)
\label{eq11}
\end{equation}

\begin{equation}
h_t = o_t * \tanh (C_t)
\label{eq12}
\end{equation}

In Equations \ref{eq11} and \ref{eq12}, the parameter $W_o$ designates the weight matrix associated with the output gate.

\subsection{ARIMA Model}\label{subsec4}
The Autoregressive Integrated Moving Average (ARIMA) model is one of the classical time series forecasting models \cite{arima1}. The ARIMA($p$, $d$, $q$) model is an extension of the ARMA($p$, $q$) model \cite{arima2}, which is built upon the ARMA model by incorporating differencing methods to transform non-stationary series into stationary series \cite{arima3}. The mathematical model of ARIMA is:

\begin{equation}
y_t = \mu + \sum_{i=1}^{p} \varphi_i \cdot y_{t-i} + \varepsilon_t + \sum_{j=1}^{q} \theta_j \varepsilon_{t-j}
\tag{2}
\end{equation}

\noindent where $\mu$ is the constant term, $\varphi_i$ is the autoregressive coefficient, $y_{t-i}$ is the lagged value, $\varepsilon_t$ is the residual, and $\theta_j$ is the moving average coefficient.

The ARIMA model expression is as follows:

\begin{equation}
\left(1 - \sum_{i=1}^{p} \varphi_i L^i\right)(1-L)^d y_t = \varphi_0 + \left(1 + \sum_{i=1}^{q} \theta_i L^i\right)\varepsilon_t
\tag{3}
\end{equation}

\noindent where $\varphi_i$ is the autoregressive coefficient, $L$ is the lag operator, $\theta_i$ is the moving average coefficient, $p$ is the autoregressive lag order, $q$ is the moving average order, and $d$ is the differencing order.

The ARIMA model establishment method is as follows:

1) Model identification and order determination. Verify the stationarity of the time series through unit root testing (ADF test). Non-stationary data need to be differenced. Use the autocorrelation function plot (ACF plot) and partial autocorrelation function plot (PACF plot) of the time series for analysis to determine the autoregressive lag order, differencing order, and moving average order.

2) Model estimation and testing. Using the data of a specific time series, estimate the model parameters and conduct tests to determine whether the model is appropriate. If not appropriate, return to the previous step \cite{intro7}.

3) Model forecasting. After the model has passed various tests, the ARIMA model can be used to forecast future time series data. The forecasting process is based on the estimated ARIMA model parameters to predict future data points.

\begin{figure}[H]
    \centering
    \includegraphics[width=12.5 cm]{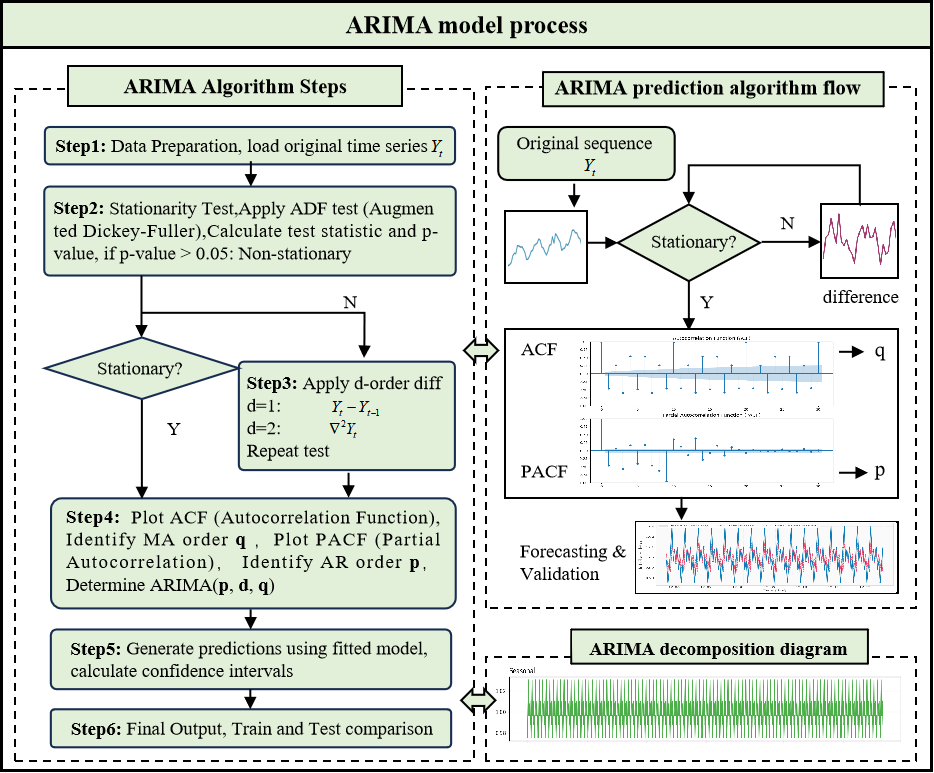}
    \caption{ARIMA Architecture Diagram.}
    \label{fig4}
\end{figure}  

The ARIMA process involves multiple steps to ensure series stationarity and optimize model parameters. As shown in Figure \ref{fig4}, first, in the data preparation stage, the original time series ($Y_i$) is loaded. Next, a stationarity test is performed using the Augmented Dickey-Fuller (ADF) test to calculate the test statistic and the $p$ value. If the $p$ value is greater than 0.05, the series is considered non-stationary and requires differencing. This differencing operation involves taking $d$-order differences (e.g., $Y_i - Y_{i-1}$ when $d=1$, and $\nabla^2 Y_i$ when $d=2$). Stationarity tests are repeated until the series is stationary. Once the series is stationary, the moving average (MA) order $q$ is identified by plotting the autocorrelation function (ACF) and the autoregressive (AR) order $p$ by plotting the partial autocorrelation function (PACF). This allows the specific parameters of the ARIMA ($p$, $d$, $q$) model to be determined. After fitting the model, forecasts are generated and confidence intervals are calculated for forecast validation. Finally, final results are output by comparing the training and test data to ensure the accuracy of the model in time series decomposition and forecasting. The whole process emphasizes the importance of stationary processing, parameter identification and verification, and reflects the systematicity and robustness of the ARIMA model in dealing with non-stationary time series.

\subsection{XGBoost Model}\label{subsec5}

As a widely adopted ensemble learning algorithm, XGBoost \cite{xgboost1} represents an advanced implementation of gradient boosting decision trees, characterized by its exceptional computational efficiency and strong scalability. The fundamental methodology underlying XGBoost involves the sequential construction of decision tree ensembles through an additive training strategy \cite{xgboost2}. This approach progressively refines predictive performance by iteratively minimizing the discrepancy between predicted and actual values, while simultaneously incorporating complexity penalties to enhance the model's ability to generalize to unseen data and maintain numerical stability \cite{xgboost3}.
  
\begin{figure}[H]
    \centering
    \includegraphics[width=12 cm]{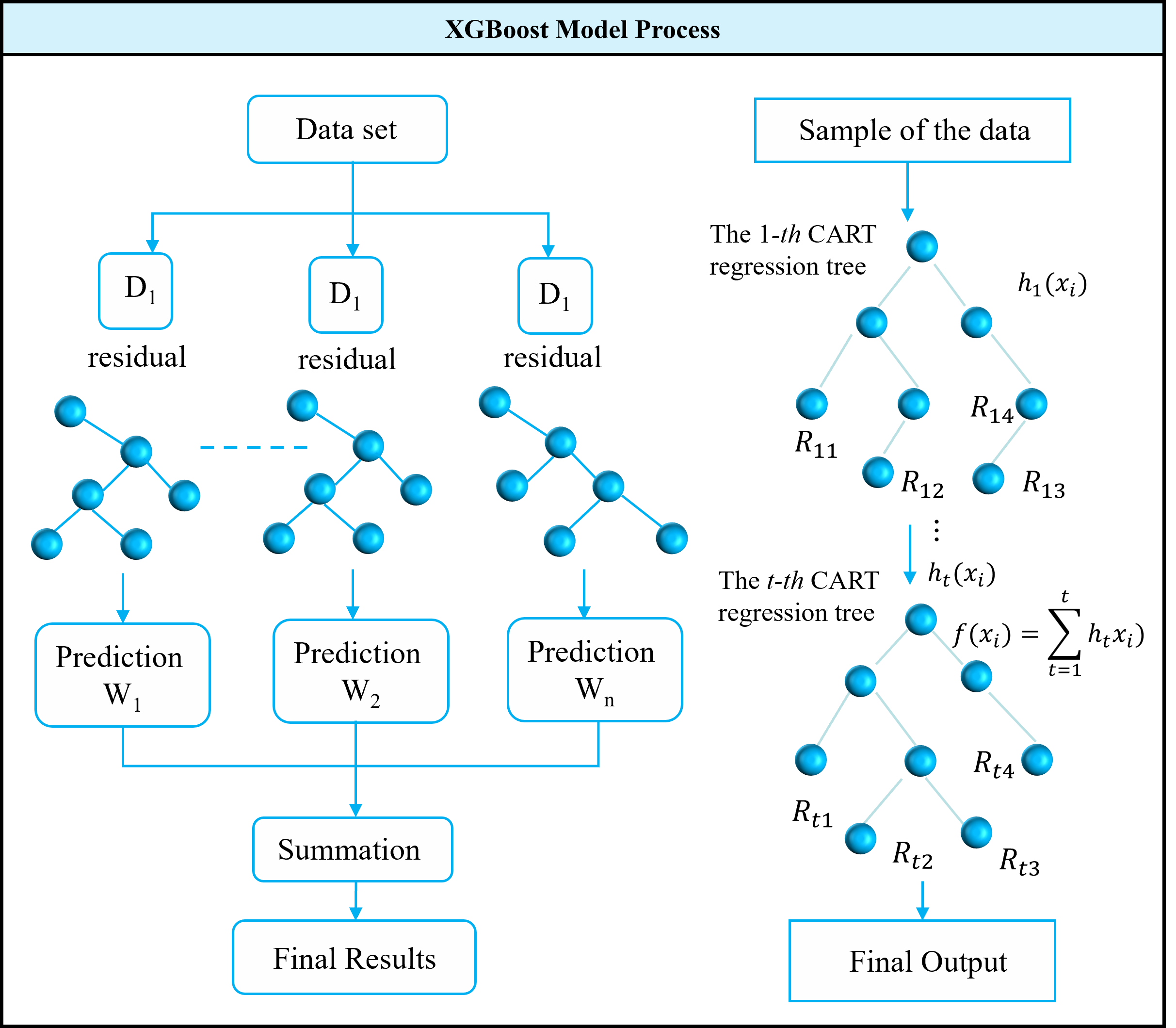}
    \caption{XGBoost Architecture Diagram.}
    \label{fig5}
\end{figure}

A distinguishing feature of XGBoost lies in its utilization of second-order Taylor polynomial approximation for the loss function during each boosting iteration. This mathematical technique enables the algorithm to capture curvature information of the objective landscape, leading to more accurate gradient estimation and substantially faster convergence compared to traditional first-order methods. The framework achieves a balance between predictive accuracy and model parsimony through a carefully designed objective function that integrates both the empirical risk and structural regularization components, The XGBoost model flow chart is shown in \ref{fig5}.

\begin{equation}
\mathcal{L}^{(t)} = \sum_{i=1}^n l(\mathbf{y}_i, \hat{\mathbf{y}}^{(t)}) + \sum_{k=1}^t \Omega(f_k)
\label{eq4}
\end{equation}

\begin{equation}
\Omega(f_k) = \gamma T + \frac{1}{2} \lambda \sum_{j=1}^{T} \omega_j^2
\label{eq5}
\end{equation}

In the formulation above, $\mathcal{L}^{(t)}$ denotes the comprehensive objective function at iteration $t$, where the first summation term $l(\mathbf{y}_i, \hat{\mathbf{y}}^{(t)})$ quantifies the prediction error for the $i$-th training instance, measuring the deviation between the ground truth $\mathbf{y}_i$ and the ensemble's prediction $\hat{\mathbf{y}}^{(t)}$ accumulated through $t$ iterations. The regularization component $\Omega(f_k)$ serves as a penalty mechanism to control the structural complexity of individual trees, where $T$ represents the total number of terminal nodes (leaves) in tree $f_k$, $\omega_j$ denotes the weight assigned to the $j$-th leaf node, and $\gamma$ and $\lambda$ are hyperparameters governing the trade-off between model complexity and fitting accuracy. Specifically, $\gamma$ controls the minimum loss reduction required to create an additional leaf partition, while $\lambda$ applies L2 regularization to leaf weights, effectively preventing overfitting by discouraging extreme weight values.
The iterative refinement process in XGBoost follows an additive model paradigm, where each newly trained tree serves to correct the residual errors of the existing ensemble. The prediction update mechanism at iteration $t$ can be expressed mathematically as:

\begin{equation}
\hat{\mathbf{y}}_i^{(t)} = \hat{\mathbf{y}}_i^{(t-1)} + \eta \cdot f_t(\mathbf{x}_i)
\label{eq6}
\end{equation}

Here, $\hat{\mathbf{y}}_i^{(t)}$ represents the cumulative prediction for sample $\mathbf{x}_i$ after incorporating $t$ trees, $\hat{\mathbf{y}}_i^{(t-1)}$ denotes the prediction from the previous iteration, and $f_t(\mathbf{x}_i)$ captures the contribution of the newly trained $t$-th tree to the prediction of instance $\mathbf{x}_i$. The learning rate parameter $\eta$ (typically $0 < \eta \leq 1$) modulates the contribution magnitude of each tree, implementing a form of shrinkage regularization that helps prevent overfitting by reducing the impact of individual trees. This additive structure embodies the essence of gradient boosting, where the algorithm progressively constructs a strong learner by combining multiple weak learners in a stage-wise fashion, with each successive tree focusing on the patterns that previous trees failed to capture. Through this iterative accumulation of tree-based predictions, XGBoost achieves remarkable predictive performance across diverse machine learning tasks while maintaining computational tractability through various algorithmic optimizations including parallel tree construction, cache-aware access patterns, and out-of-core computation capabilities.

\subsection{LSTM-ARIMA-XGBoost combination model}
\begin{figure}[H]
    \centering
    \includegraphics[width=14cm]{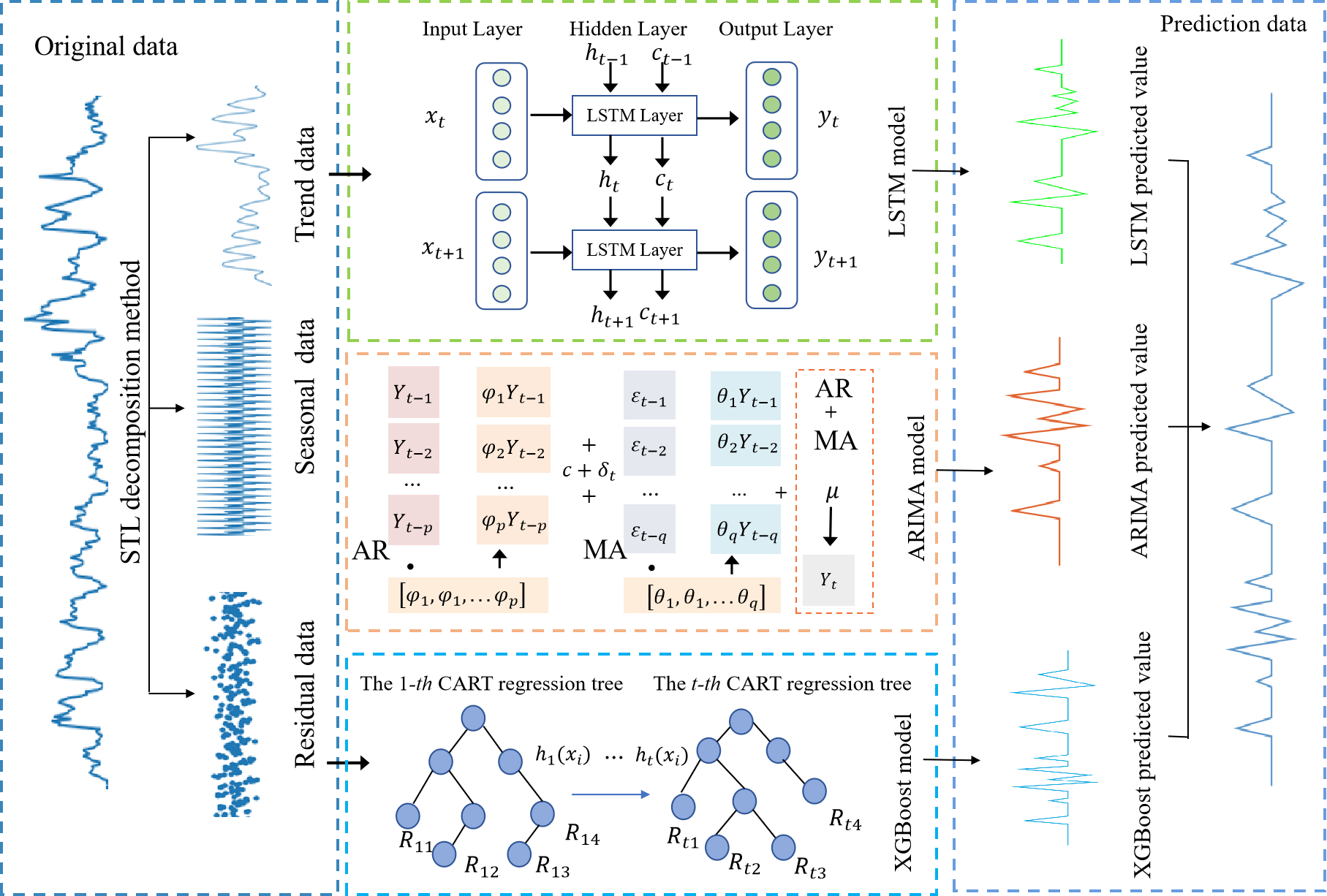}
    \caption{LSTM-ARIMA-XGBoost combination model flow chart.}
    \label{fig6}
\end{figure}

The model training dataset consists of 998 traffic flow data records from a New York City intersection between November 11, 2015, and December 12, 2015. The traffic flow data was first decomposed into trend values, residuals, and period values using the STL decomposition method. This STL decomposition method employs a multiplicative model with a defined period length of 10. The original series is first smoothed using a moving average to eliminate seasonal fluctuations and noise, capturing long-term trends and obtaining trend values. The detrended series is then segmented and averaged by period (10 time points) to generate a recurring seasonal pattern and obtain period values. Finally, the trend and period values are subtracted from the original data, leaving the remainder as the residual, representing unexplained random fluctuations. After the STL decomposition, the trend values are predicted using an LSTM model with a modified hidden layer. The data combined with the period values are predicted using an ARIMA model. Finally, the residuals are predicted using an XGBoost model. Finally, the LSTM, ARIMA, and XGBoost predictions are combined using a multiplicative model to produce the final prediction. Figure \ref{fig6} shows the flow chart of the LSTM-ARIMA-XGBoost combination model.

\section{Theoretical and experimental analysis}
\subsection{Model operating environment}
In this paper, PyCharm is selected as the development tool, Python 3.8.5 is the programming language, NVIDIA GeForce GTX4060 is selected as the graphics processing unit (GPU), i7-13600H is selected as the CPU, and 8G video memory is selected.

\subsection{Evaluation Metrics}
In this study, four representative error evaluation metrics were adopted to assess the accuracy of model predictions, namely Root Mean Square Error (RMSE), Mean Absolute Percentage Error (MAPE), Mean Absolute Error (MAE), and the coefficient of determination ($R^2$). The mathematical formulations of these evaluation indices are expressed as follows:

\begin{equation}
RMSE = \sqrt{\frac{1}{n} \sum_{i=1}^{n} (F_i - R_i)^2}
\label{eq10}
\end{equation}

\begin{equation}
MAE = \frac{1}{n} \sum_{i=1}^{n} \left| F_i - R_i \right|
\label{eq12}
\end{equation}

\begin{equation}
R^2 = 1 - \frac{\sum_{i=1}^{n} (F_i - R_i)^2}{\sum_{i=1}^{n} (R_i - \bar{A})^2}
\label{eq13}
\end{equation}

where $F_i$ and $R_i$ represent the predicted and true values of the $i$-th observation, respectively; $n$ denotes the total number of samples; and $\bar{A}$ indicates the average of all actual observations. The metrics RMSE, MAPE, and MAE quantify prediction deviations from different perspectives. Lower values of these indicators correspond to smaller prediction errors, implying higher model accuracy and more stable performance. The coefficient of determination ($R^2$), ranging between 0 and 1, evaluates the proportion of variance in the observed data that can be explained by the model. A higher $R^2$ value suggests that the model provides a better fit to the data and exhibits stronger predictive reliability.

\subsection{Original data STL decomposition}

To capture the complex temporal patterns inherent in traffic flow data, we employ a multiplicative season-trend decomposition based on Loess to decompose the observed time series into three distinct components: trend, seasonality, and residuals. The multiplicative model is used because it can handle time series where the magnitude of seasonal fluctuations varies proportionally to the trend level, a common trend in traffic flow patterns.
\begin{figure}[H]
    \centering
    \includegraphics[width=12.5 cm]{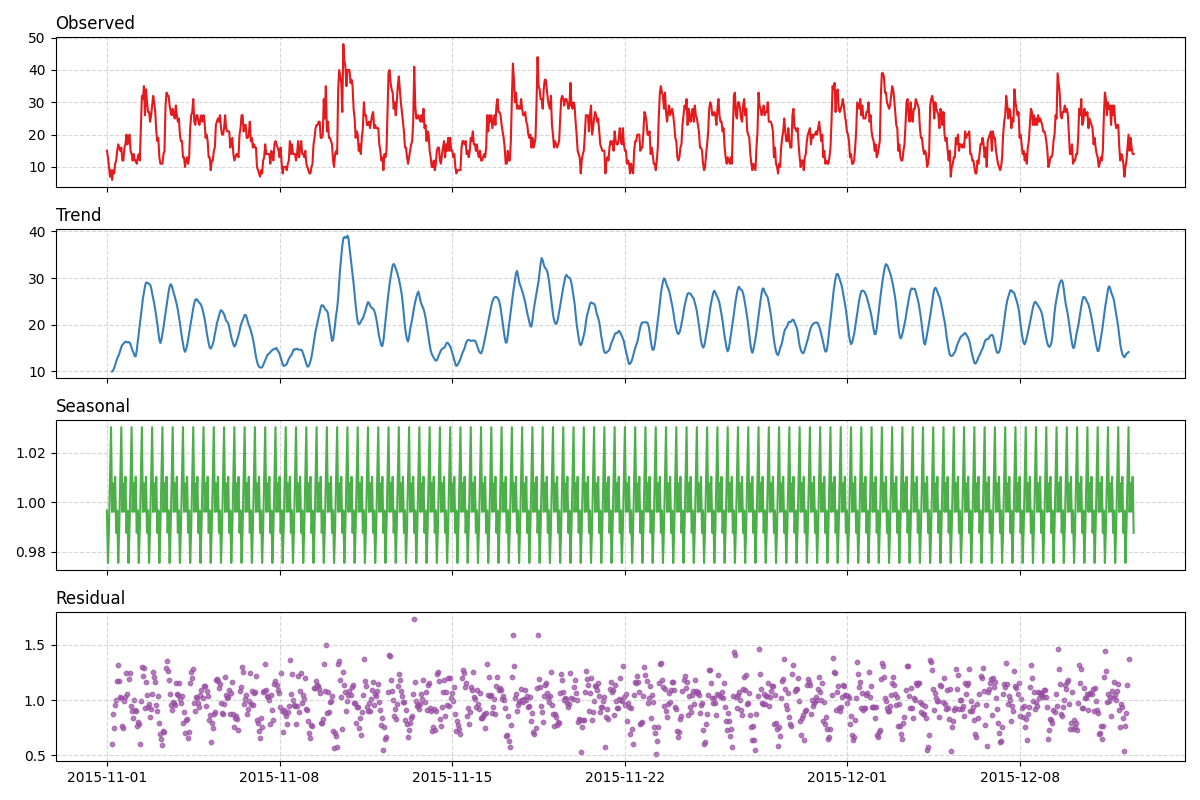}
    \caption{STL decomposition result diagram.}
    \label{fig7}
\end{figure}   

This Figure \ref{fig7} shows the STL time series decomposition of traffic flow data from November to December 2015, including observations, trend components, seasonal components, and residuals. Overall, traffic flow fluctuates between 5 and 50, with significant high-frequency fluctuations and irregular peaks, indicating that it is influenced by multiple factors. The trend component is relatively smooth, reflecting medium-term fluctuations: traffic flow peaks at approximately 40 in mid-November, then declines and recovers at the end of the month, exhibiting an overall wave-like pattern without a clear long-term upward or downward trend. The seasonal component exhibits highly regular, cyclical "sawtooth" fluctuations with an amplitude stable within ±0.02, indicating a strong and stable cyclical pattern in traffic flow, likely corresponding to periodic behavior such as weekdays and weekends, morning and evening rush hours, etc. The residuals are primarily concentrated between 0.8 and 1.4, showing a random distribution, with only a few outliers (approximately 1.6–1.7) that may be due to unexpected events or special circumstances. Overall, seasonal changes dominate the traffic flow, the trend changes are relatively gentle, and the residuals are small and random, indicating that STL decomposition can effectively extract the main time series characteristics of traffic flow data.
\subsection{LSTM modeling of trend values}
The experimental platform utilizes the Keras framework. Based on extensive experimental comparisons and multiple iterative validation tests, the parameters of the LSTM model were determined as shown in Table~\ref{tab:lstm_params}.
\begin{figure}[h]
    \centering
    \includegraphics[width=12 cm]{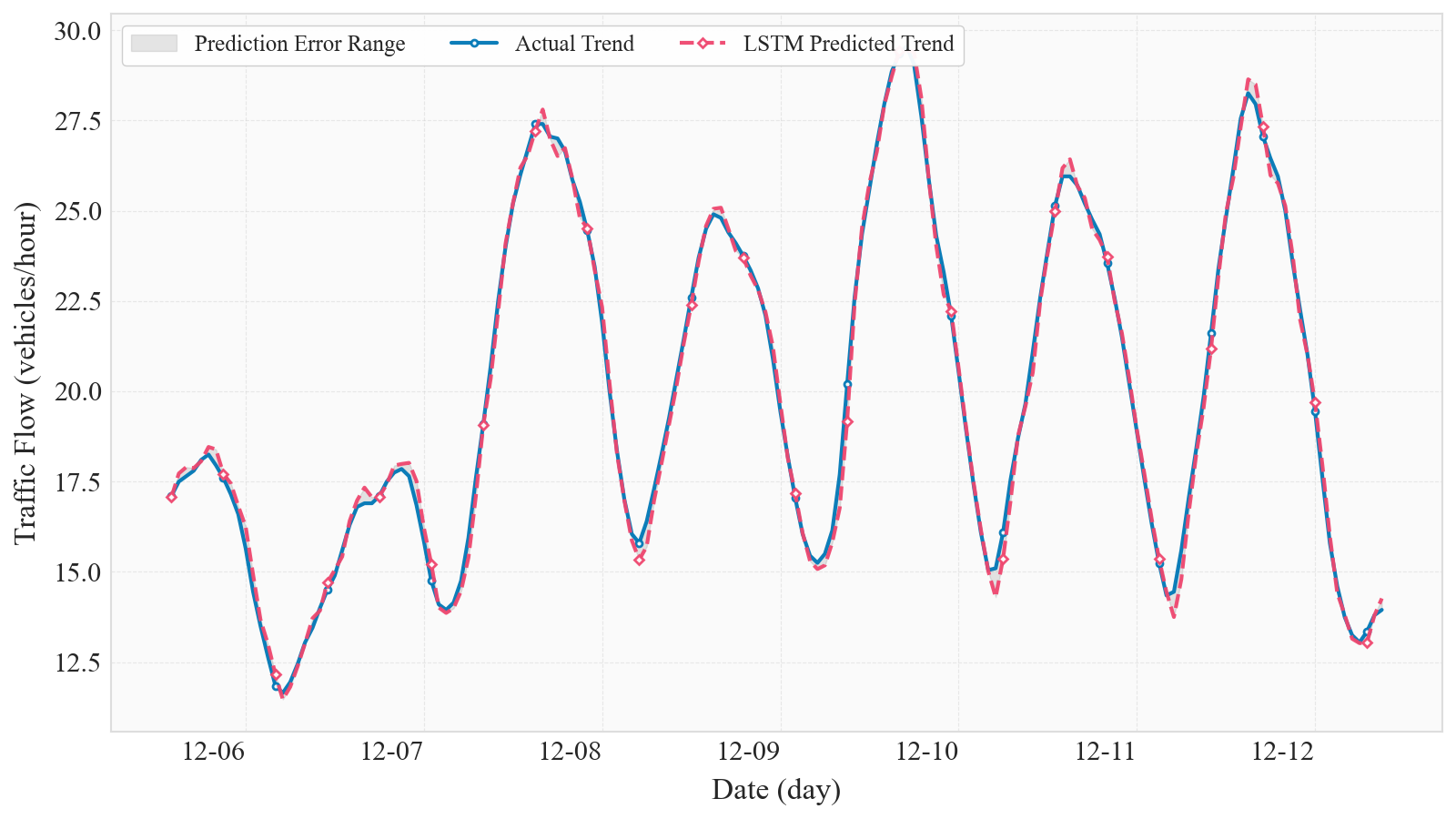}
    \caption{Comparison chart of actual trend value and predicted trend value}
    \label{fig8}
\end{figure} 
\begin{table}[htbp]
\centering
\caption{LSTM Model Parameters}
\label{tab:lstm_params}
\begin{tabular}{lc}
\hline
\textbf{Parameter Name} & \textbf{Parameter Value} \\
\hline
Learning Rate & 0.001 \\
Iterations & 200 \\
Batch Size & 32 \\
Activation Function & ReLU \\
Optimizer & Adam \\
Hidden Layer Neurons & 128 \\
Dropout & 0.2 \\
\hline
\end{tabular}
\end{table}
The dataset selected for this experiment consists of 998 data records, of which the training set accounts for 0.8 (798 records) and the testing set accounts for 0.8 (199 records). After training the LSTM model on the data, the relationship between actual trend values and predicted trend values was obtained, as shown in Figure~\ref{fig8}. In the figure, the blue line represents the original data, and the yellow line represents the predicted trend values. Based on the graphical results, it can be observed that the model fitting performance is satisfactory.

\subsection{ARIMA Modeling of Periodic Values}
\begin{figure}[h]
    \centering
    \includegraphics[width=12 cm]{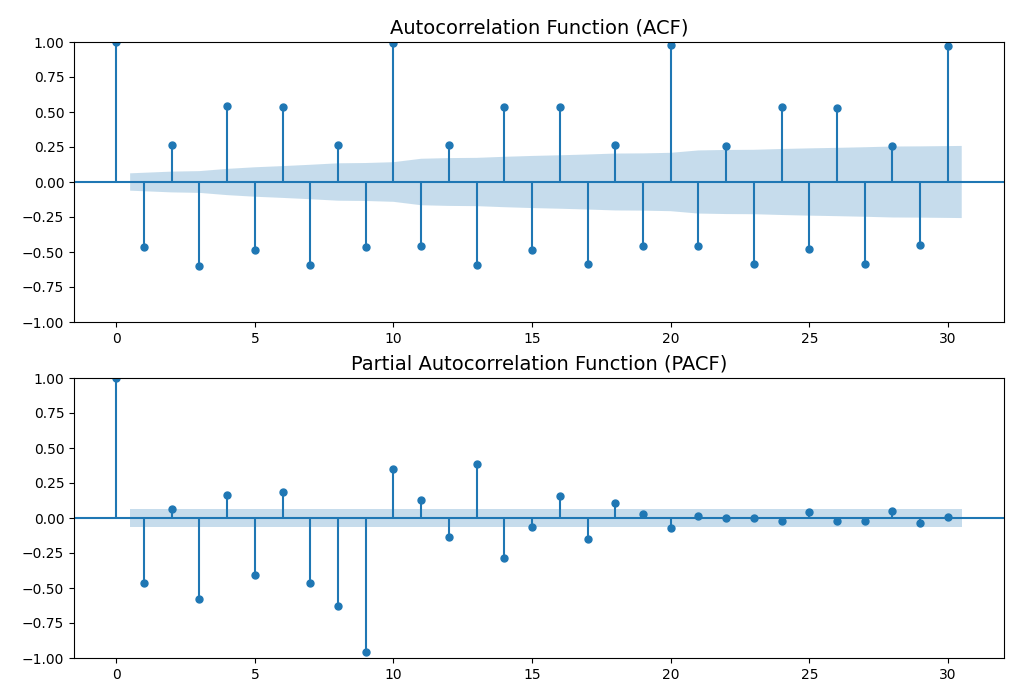}
    \caption{Comparison chart of actual trend value and predicted trend value}
    \label{fig9}
\end{figure}

To ensure the stationarity of the time series, this study first performed first-order differencing (d=1) on the original traffic flow data and verified the stationarity of the differenced series using the ADF test. As shown in Figure \ref{fig9}, the autocorrelation function (ACF) and partial autocorrelation function (PACF) plots of the differenced series were plotted to identify the model's order parameters.
The ACF test plots reveal that the autocorrelation coefficient significantly exceeds the 95\% confidence interval (blue shaded area) at several lag orders and exhibits a slowly decaying tail, accompanied by a distinct cyclical fluctuation pattern, indicating a strong autocorrelation structure. In particular, significant positive correlations are observed at lags 1, 3, 5, and 9, while significant negative correlations are observed at lags 2, 4, 6, and 8. Overall, the series exhibits alternating oscillations and slow convergence.
In the PACF test plot, the partial autocorrelation coefficient significantly exceeds the confidence interval at lags 1 to 9, with strong significance at lags 1 and 8-9. It then gradually decays and tends to truncate at lag 10, indicating that the direct correlation effect is primarily concentrated in the first nine lags. According to the Box-Jenkins methodology, the ACF exhibits a tailing pattern, while the PACF truncates after a certain number of lags, suggesting that the autoregressive moving average (ARMA) model is more appropriate.
\begin{figure}[h]
    \centering
    \includegraphics[width=12 cm]{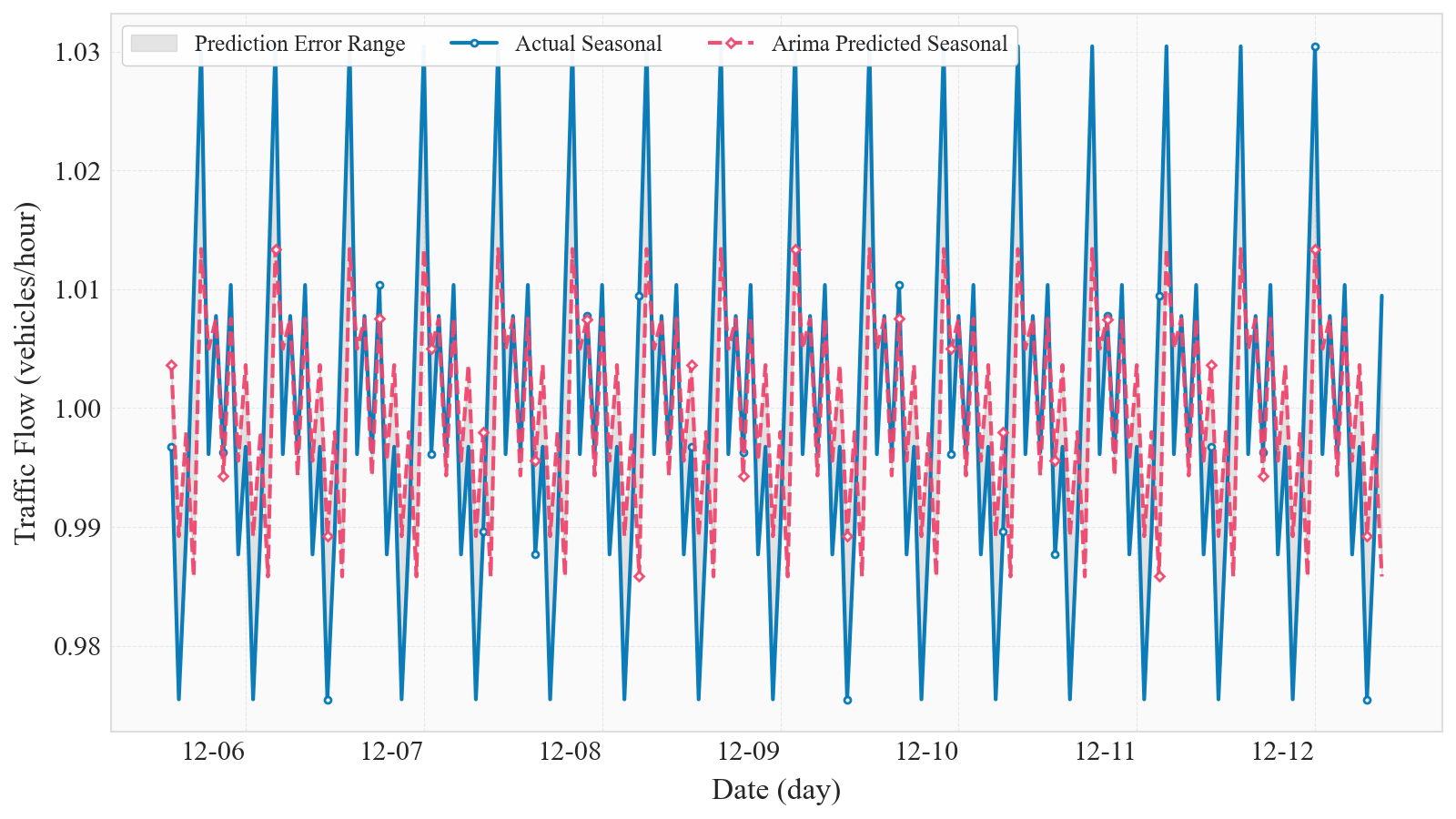}
    \caption{Comparison chart of actual trend value and predicted seasonal value}
    \label{fig10}
\end{figure} 
Combining the characteristics of the ACF and PACF, and taking into account the cyclical nature of traffic flow data, this study ultimately selected the ARIMA(2,1,2) model for fitting and forecasting. Here, p=2 represents the order of the autoregressive term, d=1 represents the first-order difference, and q=2 represents the order of the moving average term. This parameter configuration effectively captures the autocorrelation structure and short-term fluctuation characteristics of the series, providing a sound modeling foundation for traffic flow forecasting.

Figure \ref{fig10} demonstrates the ARIMA model's performance in predicting the seasonal component of traffic flow data. As can be seen from the figure, the actual seasonal component (solid blue line) exhibits a highly regular cyclical fluctuation pattern, with a fluctuation range of approximately 0.98 to 1.03 vehicles/hour, indicating a stable daily cycle in traffic flow. The red dashed line represents the seasonal component predicted by the ARIMA model, and the two curves maintain high consistency throughout the forecast period. The gray shaded area represents the forecast error range, which is extremely small, demonstrating that the ARIMA model successfully captures the cyclical patterns of the time series. The forecast curve accurately reproduces the location and amplitude of the peaks and troughs, with a period of approximately one day, which is consistent with the diurnal fluctuation pattern of actual traffic flow. Overall, the ARIMA model performs well in predicting the seasonal component, validating its strong ability to model cyclical patterns.

\subsection{XGBoost Modeling of Residual Values}
To capture the nonlinear patterns and complex dependencies in ARIMA residuals, an XGBoost (eXtreme Gradient Boosting) model was employed for  modeling. The feature engineering process incorporated three categories of variables: (1) lag features at steps 1, 2, 3, 24, 48, and 168 to capture short-term continuity, daily, and weekly periodicities; (2) rolling statistics including moving averages and standard deviations over windows of 3, 24, and 168 steps to characterize local trends and volatility; and (3) temporal features including hour-of-day, day-of-week, and weekend indicators to encode diurnal and weekly patterns. The dataset was partitioned chronologically into training (80\%) and testing (20\%) sets to preserve temporal dependencies and prevent data leakage.
\begin{figure}[h]
    \centering
    \includegraphics[width=12 cm]{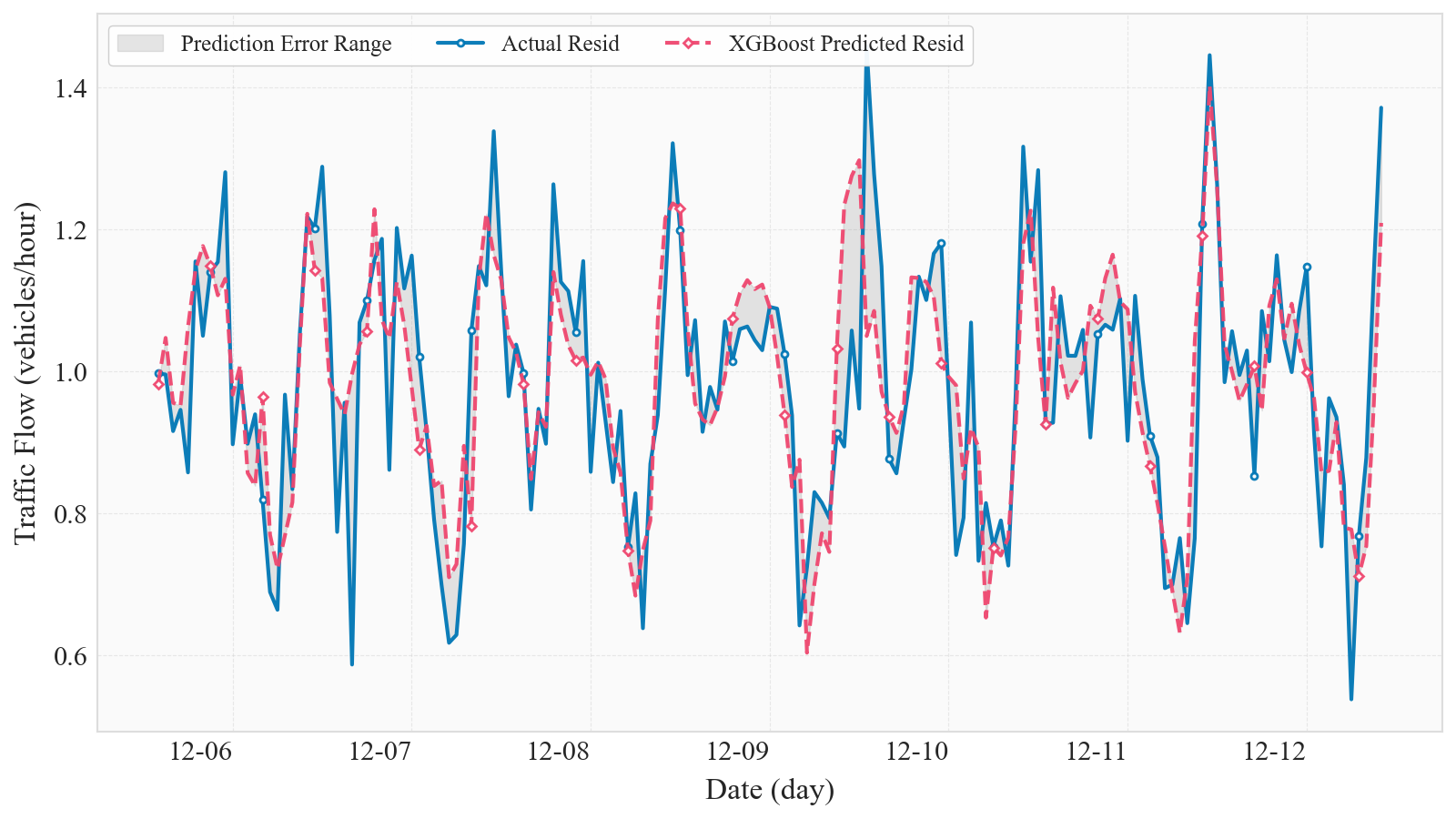}
    \caption{Comparison chart of actual trend value and predicted residual value}
    \label{fig11}
\end{figure} 
The XGBoost model was configured with the following hyperparameters: 1000 decision trees (n\_estimators), learning rate of 0.05, maximum tree depth of 6, subsample ratio of 0.8, and feature sampling ratio of 0.8 per tree. These parameters were selected to balance model complexity and generalization capability while mitigating overfitting risks. Model performance was evaluated using Mean Absolute Error (MAE), Root Mean Square Error (RMSE), and coefficient of determination ($R^2$) metrics on the test set.

Figure \ref{fig11} shows the XGBoost model's prediction performance on the residual component of traffic flow data. Compared to the seasonal component, the residual component exhibits more complex and irregular fluctuations, ranging from approximately 0.6 to 1.4 vehicles/hour, significantly larger than the seasonal component's amplitude. These irregular fluctuations reflect the impact of random disturbances and unexpected events in the traffic system that are difficult to explain with periodic patterns. As can be seen from the figure, the residual component predicted by XGBoost (dashed red line) generally follows the trend of the actual residual (solid blue line), successfully capturing most of the sudden changes and fluctuation patterns. Although there is some degree of prediction bias in some localized areas (such as the period from December 9 to December 10), the XGBoost model generally demonstrates strong nonlinear fitting capabilities. The gray-shaded area shows a relatively small range of prediction errors, indicating that the model's predictions of random fluctuations are reliable. This result validates the effectiveness of machine learning methods such as XGBoost in handling irregular and non-stationary components in time series.

\subsection{Analysis of combined model prediction results}
The LSTM-ARIMA-XGBoost combination model proposed in this paper uses the STL decomposition multiplication strategy for decomposition. Therefore, this paper multiplies the trend value predicted by LSTM, the period value predicted by ARIMA, and the residual predicted by the XGBoost model according to the time series. The comparison between the predicted value and the true value of the combination model is shown in Figure \ref{fig12} below.
\begin{table}[htbp]
\centering
\caption{Performance comparison of different models for traffic flow forecasting}
\label{tab:model_comparison}
\begin{tabular}{lccc}
\hline
Model & MAE & RMSE & $R^2$ \\
\hline
xLSTM & 2.9273 & 3.6768 & 0.6838 \\
LSTM & 2.6464 & 3.4453 & 0.7266 \\
sLSTM & 2.9112 & 3.701 & 0.6846 \\
LSTM & 2.9426 & 3.7187 & 0.6819 \\
ARIMA & 3.5157 & 4.8141 & 0.5965 \\
XGBoost & 3.4178 & 4.6486 & 0.6258 \\
xLSTM-ARIMA-XGBoost & 0.8831 & 1.0771 & 0.9742 \\
sLSTM-ARIMA-XGBoost & 0.8871 & 1.0675 & 0.9747 \\
mLSTM-ARIMA-XGBoost & 0.8782 & 1.0602 & 0.9750 \\
LSTM-ARIMA-XGBoost & \textbf{0.2959} & \textbf{0.3816} & \textbf{0.9968} \\
\hline
\end{tabular}
\end{table}
To verify the superiority of the LSTM-ARIMA-XGBoost combined model over other models such as LSTM, ARIMA, xLSTM, sLSTM, mLSTM, and XGBoost in predicting traffic flow time series data, the prediction results of the seven models were tested using the MAE, RMSE, and $R^2$ evaluation indicators. The results are shown in Table \ref{tab:model_comparison}. The table shows that the LSTM-ARIMA-XGBoost combined model outperforms other single models in all evaluation indicators and has a higher prediction accuracy.

\begin{figure}[h]
    \centering
    \includegraphics[width=13 cm]{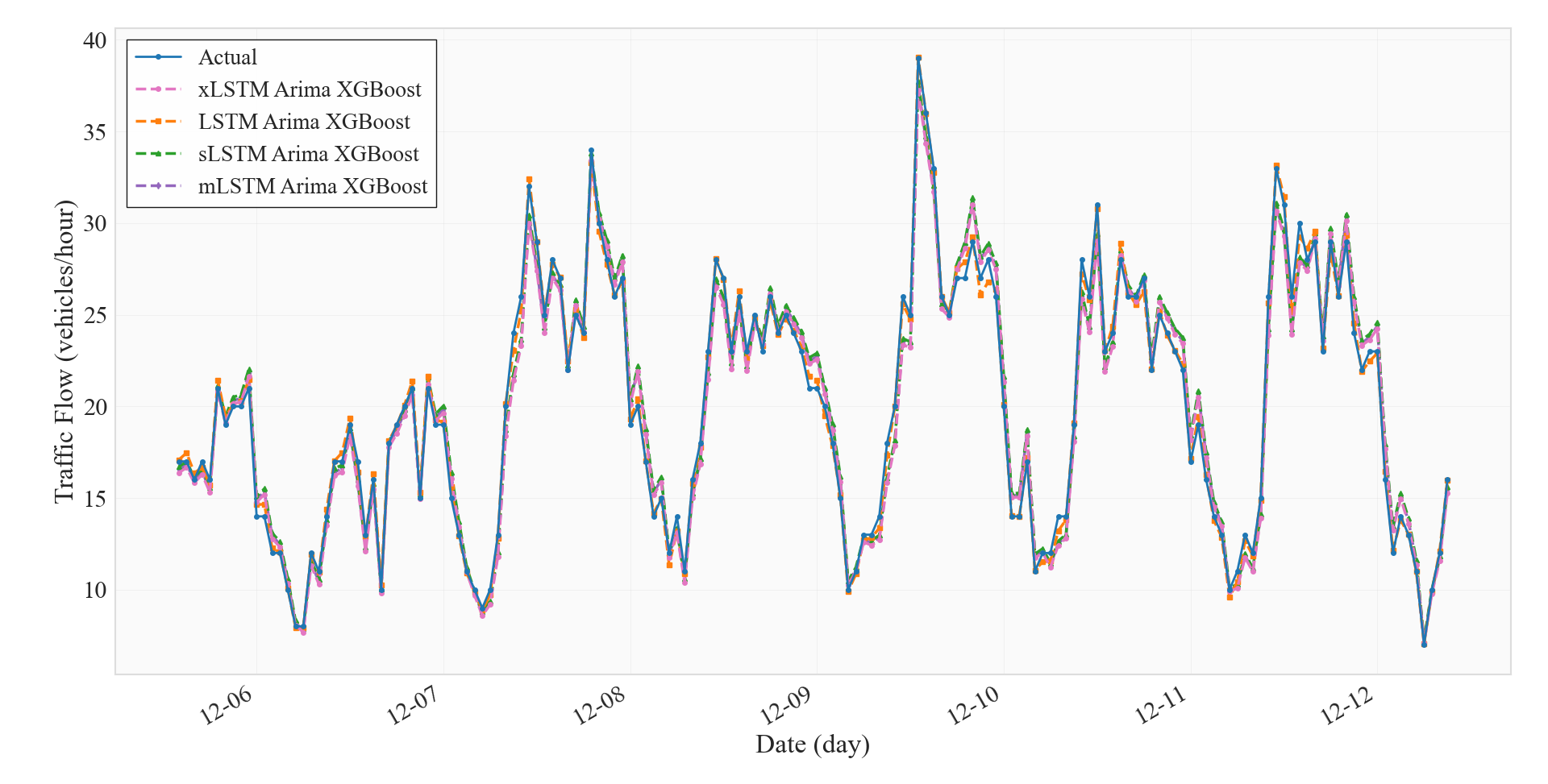}
    \caption{Comparison chart of various LSTM derivative combination models}
    \label{fig12}
\end{figure} 

\section{Conclusion}\label{sec13}
This study introduces a decomposition-driven hybrid forecasting framework that effectively addresses the challenges of accurate traffic flow prediction in complex urban environments. By leveraging the Seasonal-Trend decomposition using Loess method, the proposed approach systematically decomposes traffic flow time series into three interpretable components—trend, seasonal, and residual—each characterized by distinct temporal patterns and dynamics. The strategic assignment of specialized predictive models to each component enables the framework to exploit the complementary strengths of neural networks, statistical models, and ensemble learning methods.
The experimental validation conducted on real-world traffic data from a New York City intersection demonstrates the superior performance of the LSTM-ARIMA-XGBoost hybrid model compared to conventional single-model approaches. The results confirm that the LSTM network effectively captures nonlinear long-term trends, the ARIMA model accurately represents periodic fluctuations with stable cyclical patterns, and the XGBoost algorithm successfully models irregular residual variations that cannot be explained by deterministic components. The multiplicative integration of these sub-model predictions yields comprehensive forecasts that consistently outperform baseline methods across MAE, RMSE, and R² evaluation metrics.
Future research directions include extending the framework to multi-step ahead forecasting with adaptive prediction horizons, incorporating spatial dependencies through graph neural network architectures to enable network-wide traffic prediction, and integrating multimodal data sources including weather information, social media feeds, and calendar events to enhance contextual awareness. Furthermore, developing automated hyperparameter optimization mechanisms and exploring online learning strategies to enable continuous model adaptation in response to evolving traffic patterns represent promising avenues for improving the framework's operational efficiency and long-term reliability. The integration of explainable artificial intelligence techniques to provide transparent interpretations of model decisions would also enhance trust and facilitate adoption by transportation management agencies.

\textbf{Author Contributions} Fujiang Yuan and Yangrui Fan contributed equally to this work. Fujiang Yuan led the conceptualization of the study, developed the hybrid modeling framework, and performed software implementation and data preprocessing. Yangrui Fan was responsible for the design of experiments, model validation, and statistical analysis. Xiaohuan Bing supervised the overall research process, guided the methodological framework, and revised the manuscript critically for important intellectual content. Zhen Tian contributed to theoretical analysis, interpretation of experimental results, and technical validation. Chunhong Yuan conducted data curation, visualization, and assisted in figure preparation. Yankang Li provided support in model parameter tuning and result evaluation. All authors discussed the results, contributed to manuscript writing, and approved the final version of the paper.

\textbf{Acknowledgement}
We acknowledge support by the Open Access Publication Funds of the Göttingen University.

\textbf{Funding}
This project has no other funding support.

\textbf{Data Availability and Access} The datasets analyzed during the current study are available from the corresponding author on reasonable request.

\section*{Declarations}
\textbf{Competing Interests} The authors declare that they have no known competing financial interests or personal relationships that could have appeared to influence the work reported in this paper.

\textbf{Ethical and Informed Consent for Data Used} The data used in this paper are all publicly available and have no ethical violations.

\bibliography{sn-bibliography}

\end{document}